\documentclass{article}
\usepackage{ijcai19}

\usepackage{algpseudocode} 
\usepackage{times}
\usepackage{soul}
\usepackage{url}
\usepackage[hidelinks]{hyperref}
\usepackage[utf8]{inputenc}
\usepackage[small]{caption}
\usepackage{graphicx}
\usepackage{amsmath}
\usepackage{booktabs}
\usepackage{algorithm}
\usepackage{amssymb}
\usepackage{paralist}
\usepackage{xcolor}
\usepackage{subfigure}
\title{Accurate Robotic Pouring for Serving Drinks}

\author{Yongqiang Huang and Yu Sun
\thanks{The authors are with the Department of Computer Science and Engineering, University of South Florida, Tampa, FL 33620, USA. Email: \texttt{yongqiang@mail.usf.edu, yusun@cse.usf.edu}}%
}

\begin{document}

\maketitle
\begin{abstract}
Pouring is the second most frequently executed motion in cooking scenarios. In this work, we present our system of accurate pouring that generates the angular velocities of the source container using recurrent neural networks. We collected demonstrations of human pouring water. We made a physical system on which the velocities of the source container were generated at each time step and executed by a motor. We tested our system on pouring water from containers that are not used for training and achieved an error of as low as 4 milliliters. We also used the system to pour oil and syrup. The accuracy achieved with oil is slightly lower than but comparable with that of water. 
\end{abstract}

\section{Introduction}

In this work we focus on the task of pouring which is one of the most commonly executed tasks in people's daily lives. In fact, pouring is the second most frequently executed motion in cooking scenarios, with the first place taken by pick-and-place \cite{paulius2016}, \cite{8460200}. 

Designed to handle time sequences, recurrent neural networks (RNN) have been chosen more often for sequence generation lately \cite{graves2013}. 
RNN is capable of modeling general dynamical systems \cite{han2004,trischler2016}.
In the past, we have explored simulating pouring trajectories using RNN \cite{8206626}. This work is based on and furthers our previous work.    


Existing works that are related to accurate pouring and generalization to different containers and liquids include the following. \cite{7041426} proposes warping the point cloud of known objects to the shape of a new object, which enables pouring gel balls from one new source cup to three different receiving containers. \cite{7989307} uses deep neural network to estimate the volume of liquid in a cup from raw visual data and uses PID controllers to control the rotation of the robot arm. 
In 30 pours the average error was 38 milliliter (mL). 
\cite{DBLP:journals/corr/abs-1810-03303} uses RGB-D point cloud of the receiving cup to determine the liquid height and PID controller to control the rotating angle of the source cup. The mean error of pouring water to three different cups is 23.9 mL, 13.2mL, and 30.5mL respectively. 
\cite{chaudo2018} uses reinforcement learning to learn the policy of pouring water in simulation and tested the policy in actual robots. In the test, the poured height is estimated from RGB-D images. The algorithm averaged a 19.96mL error over 40 pours, and it generalized to milk, orange juice and apple juice but not to olive oil. 

Our work demonstrated the capability of supervised learning of learning the pouring policy for water. We obtain the poured volume from a force sensor. As straightforward as it may be, the poured volume fluctuates as the pouring proceeds, which the policy must handle.     
 
\section{Problem Description \& Approach}

\begin{figure}
\includegraphics[width=\linewidth]{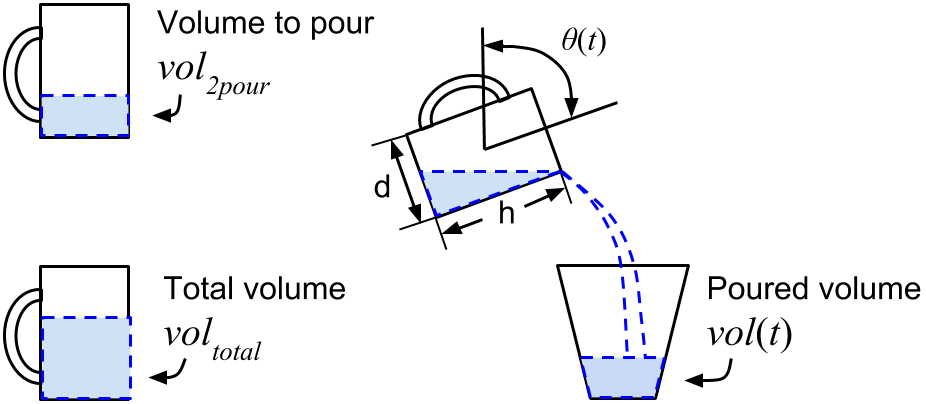}
\caption{An illustration of the six input features/dimensions for the RNN. $vol_{total}$ and $vol_{2pour}$ are the total volume and the volume to pour out. $d$ and $h$ are the diameter and height of the source container. $\theta(t)$ and $vol(t)$ are the sequences of the rotation angle and of the poured volume.}
\label{fig-pouring_scene}
\end{figure}

We define the task of accurate pouring as pouring a requested volume accurately from a source container to a receiving container. The pouring system is a complicated nonlinear time-variant system that can be affected by many factors including factors that change with time and those that are static. 

The angular velocity of the source container 
is the action that pushes the pouring process forward. To perform pouring, we need to generate the angular velocity. The velocity generator needs to take the target volume as input. It also needs to be sequential. At any time step during pouring, the generator should take the current poured volume as input, compare it with the target volume, and adjust the velocity accordingly.

We use RNN to model the velocity generator. RNN is a class of neural networks that is designed to process its inputs in order. It feeds its output from one time step into its input to the next time step. In our work, we use a variant of RNN, the peephole LSTM \cite{Gers:2003:LPT:944919.944925}. Figure \ref{fig-pouring_scene} illustrates the six input features/dimensions for the RNN.   

\section{Experiments \& Evaluation}

\begin{figure}
\includegraphics[width=\linewidth]{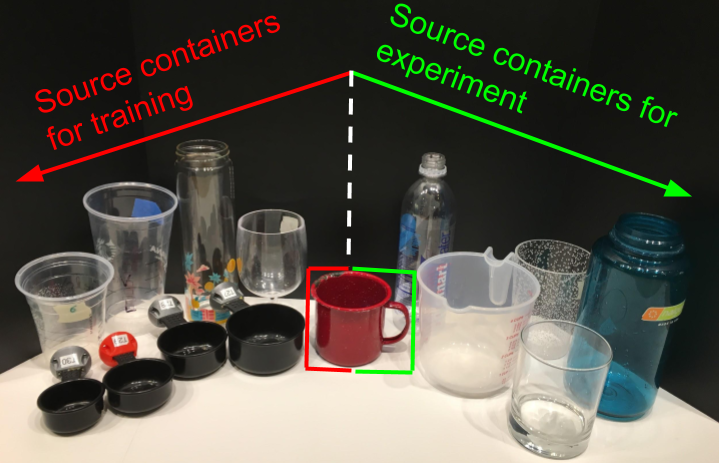}
\caption{Cups for training and evaluation}
\label{fig-cups}
\end{figure}

Data for daily interactive manipulations are available, which include some of our own \cite{huangbigdata}, \cite{doi:10.1177/0278364919849091}. We collected 284 trials of human pouring water with cups shown on the left side of Figure \ref{fig-cups}. The data are available at \url{http://rpal.cse.usf.edu/datasets_manipulation.html}. Every trial includes the six input dimensions illustrated in Fig. \ref{fig-pouring_scene}. We collected the weight of the water using a force sensor and converted it to volume. We trained RNNs with different numbers of layers and units and we found the model with 1 layer and 16 units had a simple structure and also performed well, with which we settled.

To evaluate our approach, we made a physical system that consists of the trained velocity generator, a Dynamixel MX-64 motor and the same force sensor with which we collected the data. The motor was placed at a certain height above the surface with the source container attached to it. The force sensor was placed below the receiving container.

We evaluated our system by testing it on pouring water from source containers shown on the right side of Figure \ref{fig-cups}. The difficulty of the task changes with different source containers. For source container, we use the system to pour 15 times, each time with \emph{arbitrary} $vol_{total}$ and $vol_{2pour}$ where $vol_{total}>vol_{2pour}$. 


We started with the task that has the lowest difficulty and tested the system on pouring water from a red cup that has been used for training. 
Then we increased the difficulty of the tasks and tested the system on pouring water from five different source containers that have not been used for training. 
Table \ref{table-error} summarizes the mean and standard deviation of the errors, $\mu_e$ and $\sigma_e$, in milliliters of the system pouring water from different source containers. The table is ordered in an increasing order of the error mean $\mu_e$.
Compared with the accuracy of using the red cup, the accuracy of using the five unseen source containers is lower, which is within expectation. It is worth noting that although lower than the accuracy of the red cup, the accuracy of the slender bottle 
is still high and is comparable with that of the red cup.


\begin{table}[h!]
\caption{Accuracy for pouring water from different source containers}
\label{table-error}
\begin{center}
\begin{tabular}{| c | c | c | c |}
\hline
cup & cup in training & $\mu_e$ (mL) & $\sigma_e$ (mL) \\
\hline
red & yes & 3.71 & 3.88\\

slender bottle & no & 4.12 & 4.29 \\

bubble & no & 6.77 & 5.76\\

glass & no & 7.32 & 8.24 \\

human & no & 12.37 & 9.80 \\

measuring cup & no & 11.29 & 12.82 \\

fat bottle & no & 12.35 & 8.88 \\

\hline
\end{tabular}
\end{center}
\end{table}             




     
We wanted to compare our system with human and therefore we asked four human subjects to do accurate water pouring with the red cup. We made an animation on a computer screen that shows the target volume and the real-time volume of water that has already been poured. The animation faithfully shows the fluctuation of the volume reading while pouring. The subjects were asked to look only at the animation and pour to the target volume. They were asked to pour naturally and with a single pour. Pouring too fast or too slow, or pouring multiple times were not allowed. We collected 10 trials with each subject, resulting 40 trials in total. We can see from Table \ref{table-error} that the system achieved a higher accuracy than human.

We wanted to find out if our system was able to generalize to liquid with different viscosity from water. Therefore, we tested the system on pouring cooking oil and syrup with the red cup, respectively. The red cup was used for training but the data only included it being used for pouring water. Therefore, pouring oil and syrup for the red cup is generalizing. 
Table \ref{table-viscosity} lists the accuracy of pouring liquids with different viscosities. 

\begin{table}
\caption{Accuracy of pouring liquids with different viscosities}
\label{table-viscosity}
\begin{center}
\begin{tabular}{| c | c | c | c |}
\hline
liquid & viscosity (cps) & $\mu_e$ (mL) & $\sigma_e$ (mL) \\
\hline
water & 1 & 3.71 & 3.88 \\
oil & 65 & 4.11 & 4.80 \\
syrup & 2000 & 15.66 & 3.43 \\
\hline
\end{tabular}
\end{center}
\end{table}




\section{Conclusion \& Future Work}
In this work, we presented our algorithm for accurate pouring. At every time step, the algorithm generates the angular velocity of the source container. We evaluated the algorithm using a physical system we devised. We tested the algorithm on pouring water from six different source containers and pouring oil and syrup. 
The accuracy vary for pouring with different source containers and it achieved high values for pouring with certain source containers and for pouring oil.     
To summarize, the presented results show that 
\begin{enumerate}
\item the system is able to pour accurately, and the accuracy exceeds existing pouring methods that also exhibit adaptability: \cite{7989307}, \cite{DBLP:journals/corr/abs-1810-03303} and \cite{chaudo2018}, 
\item the system is able to generalize to different source containers,
\item the system performs better than a most recent pouring algorithm based on model predictive control \cite{Chen2019IROS}
\end{enumerate}

For the future, we plan to adjust the velocity based on the difference between the trajectories of the poured volume in the data and the observed and possibly incomplete trajectory of the poured volume, in order to enhance the algorithm's generalization ability.

\bibliographystyle{named}
\bibliography{Ref_ijcai19}

\end{document}